\pdfoutput=1
\documentclass[11pt, letterpaper]{article}
\usepackage[margin=1in]{geometry} 
\usepackage{mathptmx} 
\usepackage[T1]{fontenc}

\usepackage{graphicx} %
\usepackage[sorting=none]{biblatex} %
\usepackage{amsmath, booktabs,multirow}
\usepackage{tabularx}
\usepackage{authblk}
\usepackage{amssymb}

\title{When Covariance Is Not Enough: Kernel-Based Regularization for Single-Cell Counterfactuals}

\author[1,2,$\dagger$]{Korbinian Traeuble}
\author[3]{Edward de Brouwer}
\author[3]{Archit Verma}
\author[3]{Doron Haviv}
\author[3]{Kipper Fletez-Brant}

\affil[1]{Institute of Computational Biology, German Research Center for Environmental Health, Helmholtz Zentrum München, Neuherberg, Germany}
\affil[2]{Department of Computer Science, TUM School of Computation, Information and Technology, Technical University of Munich, Garching, Germany}
\affil[3]{Computational Sciences Center of Excellence, Genentech, South San Francisco, CA, USA}
\affil[$\dagger$]{Work was done during internship at Genentech}
\date{November 2026}

\begin{document}

\maketitle

\begin{abstract}
Counterfactual editing of single cells, predicting how an observed cell would look
had one factor been different, enables \emph{in-silico} screening where experiments
are infeasible, but is useful only if the edit is faithful to the individual and not
merely to the condition. Concept bottleneck generative models make the effects of interventions
interpretable by decomposing a cell's representation into known and unknown factors. Previous work uses a linear cross-covariance penalty to disentangle known and unknown dimensions. We demonstrate that this first-order criterion does not fully separate information when non-linear interactions are present. We first demonstrate the inadequacy of linear cross-factor penalization with simulation experiments with controlled edits and known non-linear covariate effects. We then demonstrate that adding a nonlinear kernel criterion, the normalized Hilbert-Schmidt Independence Criterion, is able to recover the true intervention effects on simulation data. Finally, we apply the non-linear concept bottleneck model to a complex single-cell RNA-seq dataset and find that it improves the ability to generate counterfactual expression profiles of cells under different tissue environments.
\end{abstract}

\section{Introduction}

Counterfactual modeling asks what an observation would have looked like had one of its
underlying factors been different when other factors are held fixed. The dominant framework is Pearl's
abduction--action--prediction~\cite{pearl2009}: infer the latent state that
explains the observation, intervene on a single factor, predict the outcome. The
edit is useful only if it is at once \emph{effective}, it produces the requested
change, and \emph{faithful}, it alters nothing else. These models are particularly valuable to observational samples in biology, where high-dimensional measurements are assumed to lie in a lower dimensional space, exhaustive sampling of potential conditions is expensive and infeasible, and a sample can usually only be measured once with a single modality. Common biological counterfactual questions include what a cell's gene expression would be in a different environment, e.g. different tissue, or under a different context, e.g. alternative drug treatment. 
Single-cell RNA sequencing~\cite{macosko2015}, measuring the gene expression of individual
cells~\cite{tang2009} from a sample, provides increasing resolution into disease biology at ever increasing cell numbers and resolution~\cite{villani2017, neftel2019}. Improved throughput and decreasing costs have enabled the  production of reference atlases across tissues and diseases~\cite{tabulamuris2018, humancellatlas2017, tabulasapiens2022,
atheroatlas2025, gbmap2025}, and enabled training of latent-variable models of the cell types and states~\cite{scvi}. However, while modern foundation models
pretrained on tens of millions of cells~\cite{geneformer2023, scgpt2024} can describe cellular states, they have fallen short on valuable counterfactual prediction. Improved counterfactual models of single-cell RNA-seq data would enable the exploration of the combinatorial space of cell types, patients, treatments and doses that is infeasible at current experiment scales.

Existing models of cellular response follow two overlapping directions. The first predicts intervention effects at the
level of populations rather than of a given cell, whether by assuming a fixed additive shift in latent space~\cite{scgen, scvidr}, conditioning a decoder on the perturbation~\cite{cpa}, propagating it through a prior knowledge graph~\cite{gears}, or mapping unpaired control and treated populations onto one
another by optimal transport~\cite{cellot, cinemaot}; the transport maps do act per cell, but they match distributions rather than reconstruct a named factor. The second uses interpretable representations with biologically named
components~\cite{expimap}. Several methods combine these directions through latent-space arithmetic over disentangled factors~\cite{michigan,scdisinfact}, latent optimization with capacity-constrained unknown-attribute
embeddings~\cite{biolord}, or causally structured concept embeddings decoded by a diffusion model~\cite{causcell2025}. All emit a profile per input cell, but their evaluations do not explicitly test whether an edit preserves donor-specific identity or matches the source donor’s response under the target condition.

The concept bottleneck generative model scCBGM combines both approaches~\cite{Bottleneck} by implementing the concept bottleneck idea~\cite{cbm2020} for a generative model of cells. The model decomposes each cell's latent representation into a supervised block of observed
concepts, such as cell type, tissue or treatment, and a set of unknown factors
that account for other unrecorded variation, from donor identity to library size and batch. Counterfactual predictions for the known factors can be generated via abduction--action--prediction: encode a cell,
replace one concept with its target value, and decode from the same unknown factors.
Because the concepts are explicit, the edit can be checked rather than trusted,
which makes it a plausible route to personalized prediction. This counterfactual, however, is only valid if the two blocks are genuinely independent, such that editing a concept does not alter another concept's effect, e.g. a donor-specific signal that makes the prediction personal. To ensure disentanglement of the latent factors and concepts, scCBGM enforces independence with a penalty on the linear cross-covariance between the blocks, a first-order criterion blind to any dependence whose conditional means agree.
Kernel criteria are the known remedy in this setting: Lopez et al.\ constrained a
single-cell VAE with the $d$-variable Hilbert-Schmidt Independence Criterion to
keep the latent independent of observed nuisance factors~\cite{lopez2018}. Here
both arguments are learned, the concept head on one side and the unknown block on
the other, and the question is not invariance to a nuisance but whether an edit
made through the concept pathway stays faithful to the individual carried in the
unknowns.

We show that the linear penalty is not always enough. On a matched pair of
simulations differing in exactly one place, the cross-covariance penalty recovers
donor-specific counterfactuals when the edited condition enters as an additive
mean shift shared across donors, and fails at every penalty strength when it
instead modulates the \emph{variance} of a donor-specific signature, a regime in
which the dependence between the blocks is entirely second-order and the penalty
reports no violation. Adding a kernel criterion, the normalized
Hilbert-Schmidt Independence Criterion~\cite{HSIC}, recovers the donor-specific
edit, and the same ordering holds on the TAURUS inflammatory bowel disease
cohort~\cite{taurus}, where the added kernel penalty produces an accurate
rectum~$\to$~ileum edit.

\section{Results}

\subsection{A single-cell simulator with known donor ground truth}
\label{sec:sim-results}

We ask whether a concept bottleneck generative model edits a cell in a \emph{donor-specific} way. The original single cell concept bottleneck paper established that flipping such a concept can move cells into
the target condition \cite{Bottleneck}. A donor-specific edit, however, should capture the expression of the cells \emph{that donor} has in the target condition rather than a generic population from the target condition. To that end we implemented a simulator in which the donor effect is known by construction, so that the correct donor-specific answer is available as ground truth even for donors the model has never seen.

\begin{figure}
    \centering
    \includegraphics[width = \textwidth]{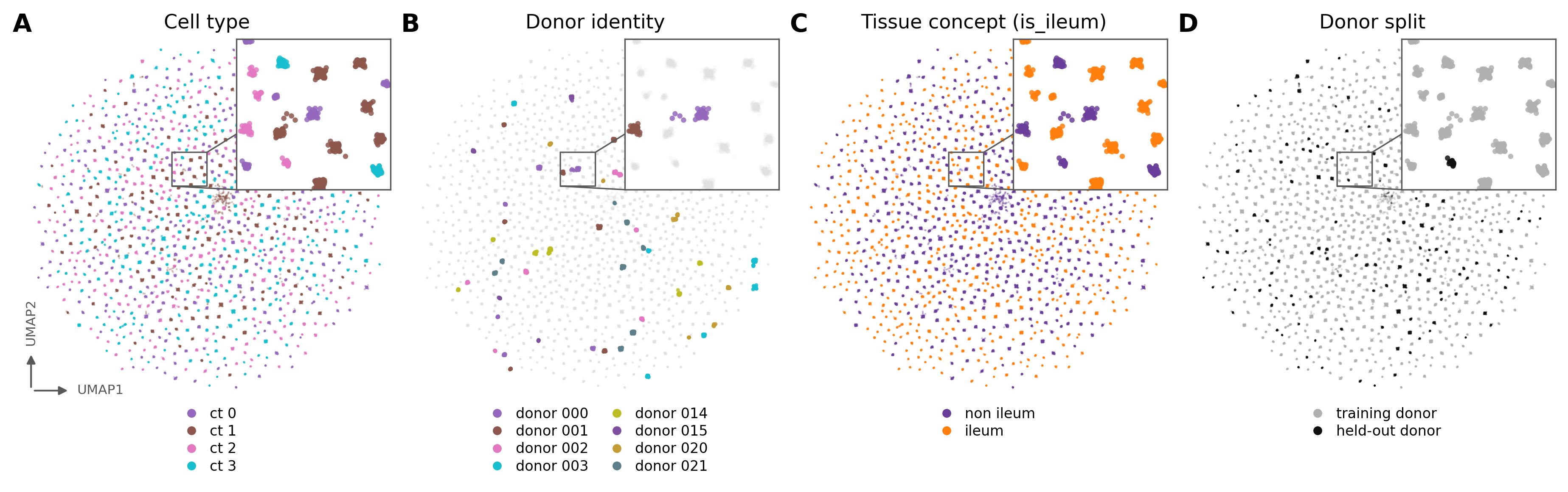}
    \caption{Structure of the synthetic donor-holdout dataset (240,000 cells, 150 donors, 20 of them held out). All panels show the same UMAP, fitted once on a uniform random subsample of 40,000 cells in the 128-dimensional PCA space the model is trained in, and recoloured: (\textbf{A}) the 4 simulated cell types; (\textbf{B}) donor identity, with 8 donors highlighted (donor\_000, donor\_001, donor\_002, donor\_003, donor\_014, donor\_015, donor\_020, donor\_021) and the remaining donors in gray; (\textbf{C}) the binary tissue concept \texttt{is\_ileum}, which is what the counterfactual flips; (\textbf{D}) training donors versus the 20 held-out donors.}
    \label{fig:sim_overview}
\end{figure}

The simulation uses a hierarchical Poisson model to generate single-cell like data \cite{Bottleneck} (Sec.~\ref{sec:sim}). Every cell draws a donor, a tissue, a cell type and a set of binary latent concepts from a fixed hierarchy. Its
Poisson log-rate is a sum of gene-wise effects: a baseline expression level, a cell-type effect, a donor fingerprint, a tissue effect, the latent concept effects, a per-cell library-size offset and gene noise (Eq.~\ref{eq:loglambda}). The two-level tissue axis plays the role of the editable binary concept \texttt{is\_ileum}. Counts are then drawn by inverse-CDF transform of stored exogenous uniforms (Eq.~\ref{eq:invcdf}), so the sampler is a deterministic function of its latent representation and we can regenerate a dataset with one effect changed and every other draw held fixed. The counts are then pushed through the pre-processing used throughout this work (median library-size normalization, $\log(1+x)$, PCA to $128$ components), so all training and
scoring happen in a $128$-dimensional PCA space. The generator produces $240{,}000$ cells over $150$ donors, $4$ cell types and $3{,}000$ genes (Table~\ref{tab:sim_hparams}). Figure~\ref{fig:sim_overview} illustrates the simulated dataset with four cell types
(A), the donor structure (B), and flipped tissue concept identity (C).

\subsection{Donor effects must be low rank for held-out donors to be predictable}

In the original single cell concept bottleneck simulator, the donor fingerprint
$\gamma_{b,g}$, the effect of donor $b$ on gene $g$, is drawn i.i.d.\ per gene, which makes each donor an isolated point in gene space: nothing learned from the training donors carries information about an unseen donor's vector $\gamma_b$, and the donor holdout is unsolvable in principle rather than merely hard. To properly model donor-specific effects, we replace the donor block by a shared factor model
$\gamma=HW$ with $H\in\mathbb{R}^{B\times R}$ and $W\in\mathbb{R}^{R\times F}$
(Eq.~\ref{eq:lowrank}), where $B=150$ is the number of donors, 
$F=3{,}000$ the number of genes, and
$R=12$ the rank: $H$ gives each donor $R$ loadings and $W$ holds $R$ gene
programs shared across donors, the analogue of a small number of transcriptional
programs that individuals up- or downregulate to different degrees rather than
$F$ independent per-gene effects. The program matrix $W$ is drawn once and
shared, and its variance is scaled by $1/R$ so that the per-gene donor variance
$\mathrm{Var}(\gamma_{b,g})=\sigma_B^2$ is unchanged, where $\sigma_B$ is the
donor-effect standard deviation: only the \emph{rank} of the donor signal
differs, not its strength. Every donor, held out or not, then lies in the same
$R$-dimensional subspace, so a donor readout learned on training donors can
transfer to unseen ones. All results below use this low-rank generator
(Table~\ref{tab:sim_hparams}).

\subsection{Linear regime: a covariance penalty is enough}
\label{sec:lin}

We simulate data with fixed donor, tissue (non-ileum and ileum), and cell type of various library sizes and noisy gene expression observations. When training the model, the cell type and tissue are observed for each cell in a supervised concept vector $\hat c$, and the unknown-factor block $u$, should account for the donor, library size and the gene noise. We test how well the model performs when generating a counterfactual by switching only the tissue concept to ileum (Sec.~\ref{sec:model}).
Donor identity is an unknown concept, so for the model to make a successful donor-specific edit the donor identity must survive in $u$ and the decoder must recombine it with the
flipped concept. What keeps $\hat c$ and $u$ from entangling is an orthogonality
penalty $\lambda_\perp\mathcal{L}_\perp(\hat c,u)$, where $\mathcal{L}_\perp$
penalizes dependence between the two blocks and $\lambda_\perp$ sets its strength;
the original model uses a linear cross-covariance penalty, and we start there.

We evaluate the model on held-out donors: $20$ of the $150$ donors are withheld from
training entirely (Figure~\ref{fig:sim_overview}D), and at inference each unseen
donor's non-ileum cells are edited to ileum and compared against that donor's observed ileal cells of the same cell type. Absolute distances are not comparable across cell types, so we score each held-out (donor $b$, cell type
$k$) pair by a ratio of squared Maximum Mean Discrepancies (MMD),
\begin{equation}
  \mathrm{rMMD}_{b,k}=
  \frac{\mathrm{MMD}\big(\hat{X}_{b,k},\,X^{\mathrm{tgt}}_{b,k}\big)}
       {\mathrm{MMD}\big(X^{\mathrm{tgt}}_{\cdot,k},\,X^{\mathrm{tgt}}_{b,k}\big)},
  \label{eq:rmmd_donor}
\end{equation}
where $\hat{X}_{b,k}$ is the edit of donor $b$'s type-$k$ source cells,
$X^{\mathrm{tgt}}_{b,k}$ that donor's own real target cells of type $k$, and
$X^{\mathrm{tgt}}_{\cdot,k}$ the generic target-condition population of type $k$
pooled over all training donors (Sec.~\ref{sec:eval}). The denominator is thus
the best "get the condition right but ignore the donor" answer: a value near or
above $1$ means a generic target-condition cell, and a value below $1$ means the
edit is closer to \emph{that donor's} target than the pooled population is, which
is exactly the donor-specific property. We report the median over all held-out
(donor, cell type) rows and, where stated, the fraction below $1$.

\begin{figure}
    \centering
    \includegraphics[width = \textwidth]{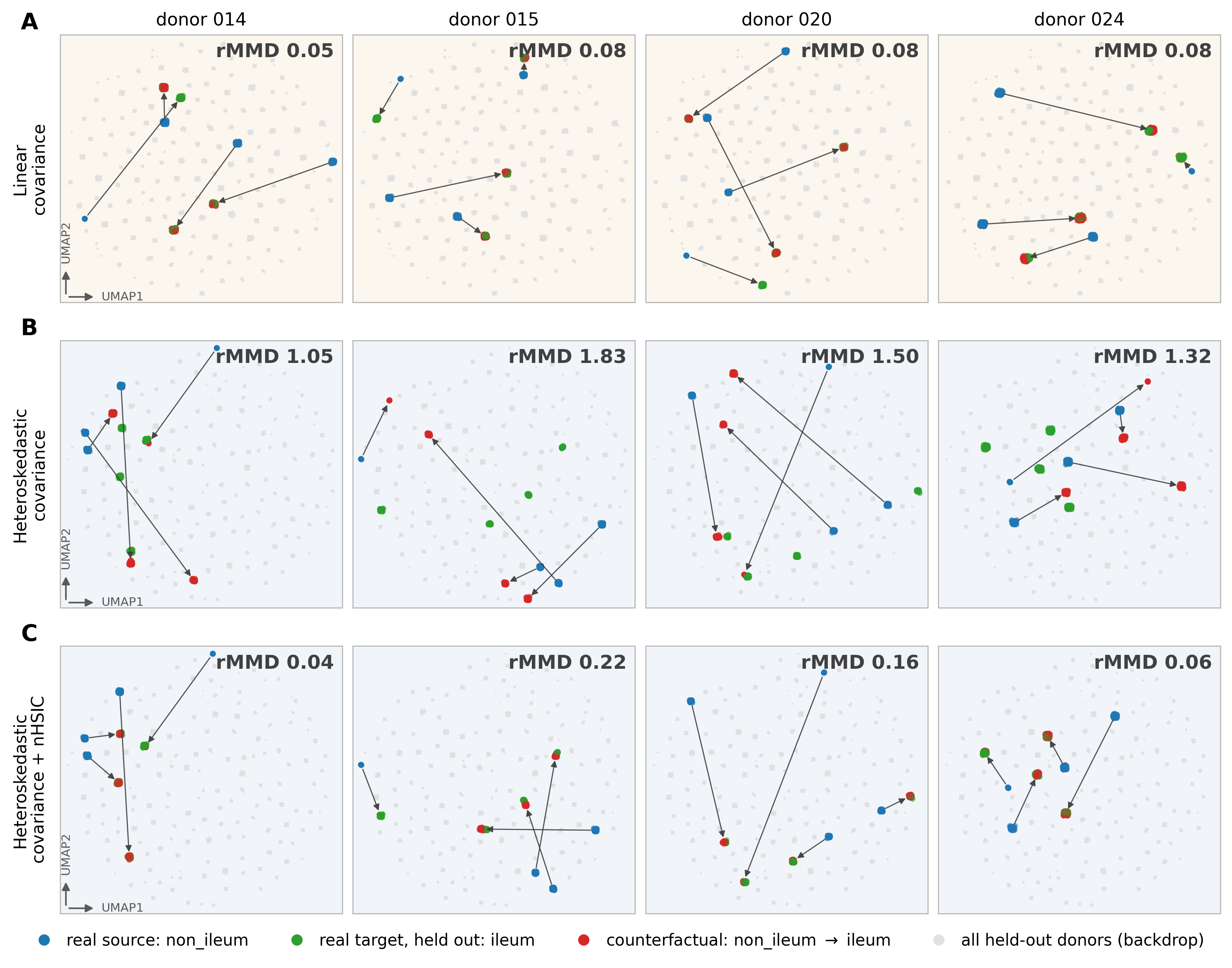}
    \caption{Donor-holdout counterfactual on the matched linear / heteroskedastic simulation pair. Rows: (A) linear regime with a covariance penalty, (B) heteroskedastic regime with the same covariance penalty, (C) heteroskedastic regime with covariance $+$ nHSIC; all at $\lambda_\perp=0.5$. Columns are held-out donors (donor\_014, donor\_015, donor\_020, donor\_024), never seen in training. Each panel shows all 4 simulated cell types of that donor: the retained \texttt{is\_ileum}$=$non\_ileum cells (blue), the withheld real \texttt{is\_ileum}$=$ileum cells (green), and the model's counterfactual edit of the blue cells (red). Rows B and C share one UMAP, fitted jointly on their real cells and both predictions, so they are directly comparable and only the red cloud moves; row A is a different dataset and therefore a different coordinate system. Each group is subsampled to at most 300 cells per donor and cell type, so the three groups are drawn at comparable density. The three groups are drawn in a single shuffled scatter, so neither is systematically painted over the other and a true overlap reads as a blend. Arrows run from each cell type's real source cloud to the model's counterfactual for that cloud (centroids), i.e. the edit the model actually made; the edit succeeds when the arrow head lands on the green held-out target. They are a reading guide in UMAP coordinates, not a displacement in any metric space. The gray backdrop is the real cells of all 20 held-out donors (20,000 of 32,769 drawn), so each gray blob is another donor. Panels are framed, and tinted by regime -- rows B and C share both an embedding and a tint, row A neither. Panel labels give that donor's median rMMD over the cell types scored for it (a cell type is scored only where the donor has enough cells on both sides); run-level medians over all 20 held-out donors: (A) 0.09, (B) 1.30, (C) 0.15 (lower is better; $<1$ beats mapping the cells to the closest existing population).}
    \label{fig:sim_cf_panel}
\end{figure}

We first evaluate a simulation where the donor-specific effect is an additive mean shift shared by all donors, so
the donor-specific part of the answer is entirely the donor's own baseline, which
the model must carry in $u$. Here, the original concept bottleneck model with a linear orthogonality penalty generalizes to unseen donors
(Table~\ref{tab:sim_rmmd}, top block): the cross-covariance penalty gives a
median held-out-donor rMMD of $0.09$, $0.06$ and $0.06$ at $\lambda_\perp=0.5$,
$1$ and $5$, all well below $1$. Row A of Figure~\ref{fig:sim_cf_panel} shows
this per donor: for four donors never seen in training, the edited cells (red)
land on the withheld real ileal cells (green) for all four cell types, and the
arrows track the real source-to-target displacement. This is the regime in which
the method behaves as advertised.

\begin{table}[t]
  \centering
  \begin{tabular}{llccc}
    \toprule
    Regime & Orth.\ loss & $\lambda_\perp{=}0.5$ & $\lambda_\perp{=}1$ & $\lambda_\perp{=}5$ \\
    \midrule
    \multirow{3}{*}{Linear (mean shift)}
      & cov & \textbf{0.09} & \textbf{0.06} & \textbf{0.06} \\
      & nHSIC & \textbf{0.10} & \textbf{0.08} & \textbf{0.06} \\
      & cov\,+\,nHSIC & \textbf{0.03} & \textbf{0.24} & 2.02 \\
    \midrule
    \multirow{3}{*}{Nonlinear (heteroskedastic)}
      & cov & 1.30 & 1.30 & 1.09 \\
      & nHSIC & 1.09 & \textbf{0.69} & \textbf{0.35} \\
      & cov\,+\,nHSIC & \textbf{0.15} & \textbf{0.21} & 2.16 \\
    \bottomrule
  \end{tabular}
  \caption{Median per-donor rMMD on \emph{held-out donors} for the matched linear/heteroskedastic simulation pair. Bold $=$ beats the baseline ($<1$). A covariance penalty suffices in the linear regime but fails at every strength in the heteroskedastic one, where the condition enters through a conditional variance; adding nHSIC recovers a donor-specific edit.}
  \label{tab:sim_rmmd}
\end{table}

\subsection{Heteroskedastic regime: a covariance penalty fails, nHSIC recovers the edit}

A general linear additive shift is an unrealistic assumption for true single-cell RNA-seq observations. We therefore build a matched
member of the pair, generated from the same seed and differing only on a block
$V$ of the $150$ highest-expression genes. On $V$ the tissue main effect is
removed and replaced by a donor--tissue variance interaction
(Eq.~\ref{eq:hetero}): $\varphi_b$ is a per-donor signature on $V$ (drawn once,
zero-mean across donors), $m(t_n)$ is a tissue multiplier equal to $1$ in
non-ileum and $M$ in ileum, $\rho=0.5$ scales the channel and $M=3$ is the ileum
variance multiplier. Each donor thus carries its own signature $\varphi_b$ whose amplitude grows by $M$ in the ileum. Because
$\mathbb{E}[\varphi]=0$, the condition and the block are uncorrelated,
$\mathrm{Cov}(\texttt{is\_ileum},V)=0$, while the conditional variance ratio
between the two tissues is $M^2$. The correct edit is to amplify each donor's own
baseline signature, and the information needed is present in the donor's source
cells, so the task stays well posed.

In this regime the covariance penalty fails at every strength
(Table~\ref{tab:sim_rmmd}, bottom block): median rMMD $1.30$, $1.30$ and $1.09$,
with only $35\%$, $37\%$ and $46\%$ of the individual (donor, cell type) rows
below $1$. Row B of Figure~\ref{fig:sim_cf_panel} shows the failure mode. The
edit is not absent, the red cloud does move away from blue, but it moves toward
a generic ileal position and does not reach the donor's own green cells.

By contrast, adding a kernel penalty \cite{HSIC} recovers the edit. The normalized
Hilbert-Schmidt Independence Criterion (nHSIC) measures the dependence between
$\hat c$ and $u$ through kernels rather than covariance, so it registers the
higher-order dependence that a covariance penalty is blind to
(Sec.~\ref{sec:nhsic}). With nHSIC alone
the median rMMD falls from $1.09$ to $0.69$ to $0.35$ as $\lambda_\perp$ goes
from $0.5$ to $1$ to $5$; combining the two penalties is better still, $0.15$ at
$\lambda_\perp=0.5$ and $0.21$ at $\lambda_\perp=1$, with $95\%$ of the
individual donor rows below $1$ in both cases. Row C of
Figure~\ref{fig:sim_cf_panel} is the same four held-out donors in the same
embedding as row B, and the only thing that changes is that the red cloud now
lands on green. Since rows B and C share a UMAP, the comparison is direct.

We find that success is dependent on the strength on the penalty. At $\lambda_\perp=5$ the
combined penalty breaks down in both regimes, rMMD $2.16$ in the heteroskedastic
member and $2.02$ in the linear one.
This is not an inert decoder. The \emph{edit ratio} measures how far the
decoder's output moves when the concept is flipped from source to target,
relative to the source reconstruction (Sec.~\ref{sec:model}); within these three
runs it grows monotonically with $\lambda_\perp$ ($0.05$, $0.08$, $0.90$). The
model is making a large edit that leaves the target
population, i.e.\ over-regularizing the bottleneck drives the prediction off the
data manifold rather than suppressing the edit. The useful range is therefore
bounded on both sides, and the two failure modes look nothing alike.

\subsection{Real data: held-out subtype counterfactual in the Taurus cohort}
Finally, we evaluate linear and non-linear penalties in concept bottleneck models on the
TAURUS cohort \cite{taurus}.  We use the pooled subtype holdout used by the original authors: all
cells of one fine subtype in the target condition are withheld from training,
the retained cells of that subtype in the source condition are edited, and the
prediction is compared to the withheld cells. The subtype was chosen by a
model-free screen of all $36$ location-agnostic subtypes, ranking them by the
real source-versus-target distance and by the number of donors that carry the
subtype on both sides (Sec.~\ref{sec:eval}), since a subtype whose two
conditions barely differ makes the ratio metric uninformative. We selected
\texttt{stem\_LGR5pos} (per-donor real gap MMD $0.20$, $14$ donors on both
sides). The holdout removes all $4{,}100$ ileal \texttt{stem\_LGR5pos} cells of
$987{,}743$; at inference the $6{,}342$ retained rectal cells of that subtype are
edited non\_ileum $\to$ ileum.

We compare three runs that differ only in the orthogonality penalty and its
weight (Table~\ref{tab:taurus_rmmd}). The covariance penalty at $\lambda_\perp=0.5$ does not produce an edit. The decoder barely responds to the flip (edit ratio $0.04$), and the prediction stays essentially where the source cells were: its MMD to the withheld ileal cells is $0.231$, against $0.244$ for the un-edited epithelial non-ileum population, which is the closest of the $14$ candidates and hence the denominator of the ratio. Read alone, the ratios (rMMD $0.95$, rSinkhorn $0.50$) are both below $1$ and would suggest success; they are below $1$ only because the source population is itself the best available answer. Raising the same penalty to $\lambda_\perp=10$ does not fix it (numerator $0.234$, rMMD $0.96$, rSinkhorn $0.46$), although the decoder does now respond to the concept (edit ratio $0.22$).

\begin{table}[t]
  \centering
  \begin{tabular}{lccc}
    \toprule
    Orthogonality penalty & $\lambda_\perp$ & rMMD $\downarrow$ & rSinkhorn $\downarrow$ \\
    \midrule
    Covariance & $0.5$ & 0.95 & 0.50 \\
    Covariance & $10$ & 0.96 & 0.46 \\
    Covariance\,+\,nHSIC & $10$ & \textbf{0.41} & \textbf{0.14} \\
    \bottomrule
  \end{tabular}
  \caption{Held-out counterfactual accuracy on the \texttt{taurus} stem holdout (Rectum $\to$ terminal ileum, \texttt{stem\_LGR5pos}). rMMD is the ratio metric of Eq.~\ref{eq:rmmd}: $<1$ means the edited cells are closer to the real held-out ileal cells than the closest existing cell population is. All three runs are below $1$, but the covariance-only numerators ($0.231$ and $0.234$) are essentially the denominator ($0.244$), i.e.\ the prediction is no closer to the target than the un-edited source it was made from; only cov\,+\,nHSIC (bold) makes a complete edit (edit ratio $0.67$ vs.\ $0.04$ and $0.22$).}
  \label{tab:taurus_rmmd}
\end{table}

\begin{figure}
    \centering
    \includegraphics[width = \textwidth]{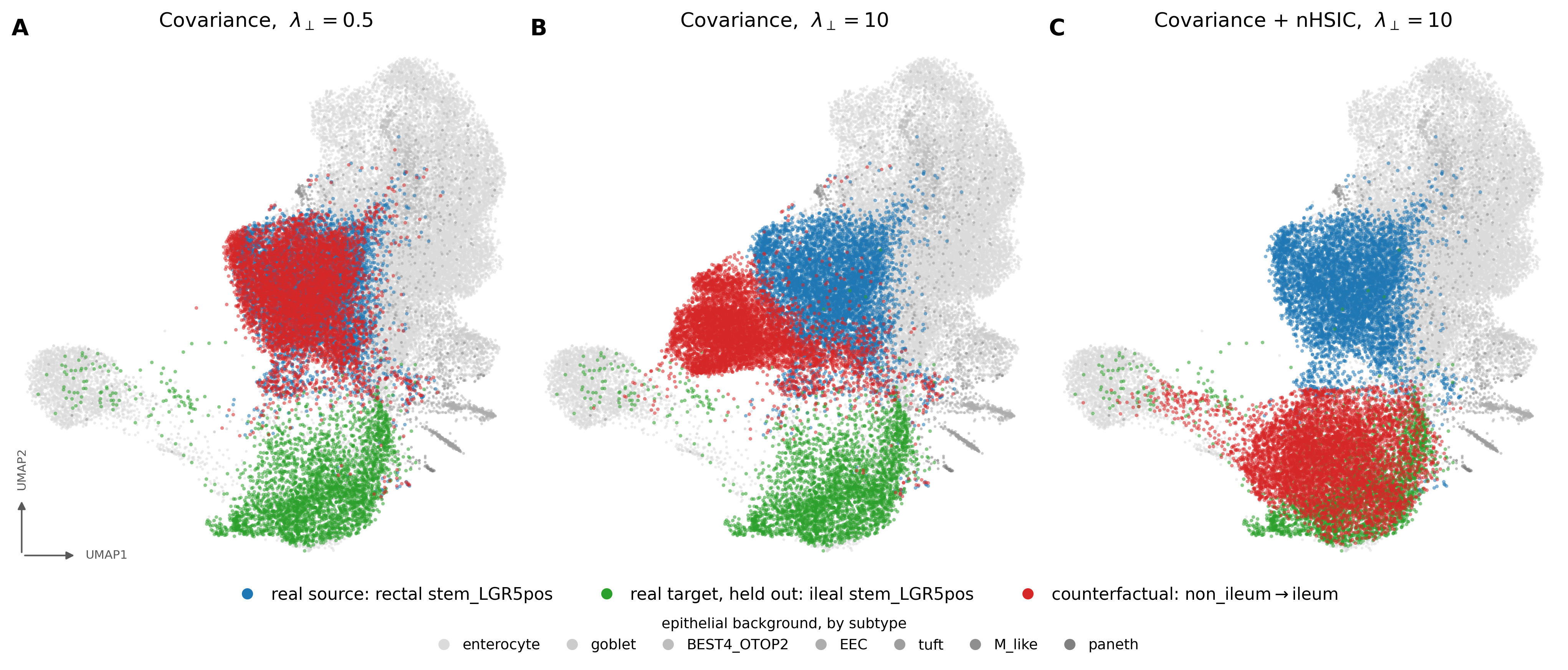}
    \caption{Counterfactual edit of the held-out \texttt{stem\_LGR5pos} subtype (rectum $\to$ terminal ileum) under three orthogonality penalties. Every \texttt{is\_ileum}$=$ileum \texttt{stem\_LGR5pos} cell ($n$=4{,}100, green) was withheld from training; at inference the retained rectal cells of the same subtype ($n$=6{,}342, blue) are edited to ileum (red) and compared against them. The gray backdrop is the rest of the \texttt{epithelial} lineage, shaded by fine subtype (largest subtype lightest); a uniform random sample of 30,000 of the 317,976 cells is drawn, to keep the UMAP tractable; sampling is uniform, so subtype proportions are preserved.}
\label{fig:taurus_cf_panel_perpanel}
\end{figure}

The combined covariance and nHSIC penalty at $\lambda_\perp=10$ gives rMMD $0.41$
and rSinkhorn $0.14$, a numerator of $0.100$ against the same denominator of
$0.244$, and an edit ratio of $0.67$.
Figure~\ref{fig:taurus_cf_panel_perpanel} shows the three runs in a shared
embedding and reproduces the ordering seen in the simulation: under
cov $\lambda_\perp=0.5$ the prediction sits on top of the source cells, under
cov $\lambda_\perp=10$ it is displaced but stops short of the withheld ileal
cells, and under cov $+$ nHSIC it overlaps them.

We find that the qualitative ranking found in the heteroskedastic
simulation, where a linear orthogonality penalty leaves an incomplete edit and a
kernel criterion completes it, is consistent in the real data. The success metric of rMMD $<1$ is
necessary but not sufficient: all three runs clear that bar, but only the run
with the kernel penalty clears it by a margin that corresponds to an actual
edit. We therefore read the ratio together with the numerator, the winning
denominator population and the decoder edit ratio.

\section{Discussion}

We asked whether a concept bottleneck generative model edits a cell in a
\emph{donor-specific} way, and the answer depends on how the edited condition
enters the data. On a matched pair of simulations with linear and non-linear donor specific effects,
the linear cross-covariance penalty of scCBGM~\cite{Bottleneck} suffices when the
condition is an additive mean shift shared by all donors and fails when it instead modulates the \emph{variance} of a donor-specific signature, regardless of the strength of the linear orthogonality penalty. The failure is not
an absent edit with a high enough penalty: the decoder moves the cells, but to a generic target-condition
position rather than that donor's own. Adding a kernel criterion, nHSIC, to the
linear one recovers the donor-specific edit, and the same holds on
Taurus~\cite{taurus}, where the covariance-only numerators sit
at the denominator while cov\,+\,nHSIC completes the edit. Neither regime is a strawman. The additive one is plausibly what the
interferon-$\beta$ stimulation benchmark of Kang et al.~\cite{kang2018} provides,
on which scCBGM was evaluated with the covariance penalty alone: a coordinated
up-regulation of an interferon program largely shared across cells and donors,
hence a first-order effect that a first-order criterion can keep out of $U$. The
heteroskedastic regime is what one expects when a perturbation amplifies an
individual's existing program instead of adding a fixed offset, as when
inflammation exaggerates a donor's baseline response. The
low-rank donor structure is likewise biologically grounded, individuals differing
mainly by up- or downregulating a limited set of shared transcriptional programs
rather than through $F$ independent per-gene effects, and it is what makes an
unseen donor predictable at all. We therefore recommend \emph{adding} the kernel penalty rather than replacing the
linear one. The terms are complementary: cross-covariance is a cheap, low-variance
estimate of the first-order dependence, whereas nHSIC must estimate a full kernel
dependence from a few hundred cells, so keeping both anchors the first order and
leaves the kernel term only what covariance cannot see. The combination reaches the
best value in both regimes at the smallest weight ($0.03$ and $0.15$ at
$\lambda_\perp=0.5$), where nHSIC alone needs $\lambda_\perp=5$ to reach $0.35$.

Hyperparameter sensitivity is the main current weakness of nonlinear concept bottleneck models. We find that over-regularization is a potential issue with the combined kernel and linear penalty. Tuning on rMMD alone would miss this, and we find that joint evaluation with the edit ratio and UMAPs is required. The best setting differed
between our two datasets ($\lambda_\perp=0.5$ in simulation, $10$ on Taurus). We did not explore the estimator's batch-size dependence and its
median-heuristic bandwidths. Our results are also single-seed, reported without intervals, 
live in a $128$-dimensional PCA space under a Gaussian loss rather than in count
space, and rest on one simulator and one cohort with a single held-out subtype.

Growing longitudinal and pan-tissue collection will eventually make further evaluations and development feasible. As data increases, we can 
move towards patient, rather than cell level, prediction and generation - treating each patient as
a distribution over cells and the cohort as a distribution over those
distributions, simultaneously generating cell type abundances and cell states. With such a readout the applications follow from the counterfactual framing, \emph{in-silico} treatment screening in a specific
patient, imputing a tissue that was never biopsied, mapping an individual along an
axis such as age, and transferring a biomarker from accessible blood to the tissue
where the disease sits. Each is an edit that must be faithful to the individual and
not merely to the condition, and which independence criterion is used decides
whether it is.

\section{Methods}

\subsection{Single cell RNA sequencing preprocessing}
Raw single cell counts from the Taurus \cite{taurus} dataset object \texttt{TAURUS\_raw\_counts\_annotated\_final.h5ad} were downloaded from \url{https://zenodo.org/records/14007626} and reduced to the 3,000 most highly-variable genes (\texttt{sc.highly\_variable\_genes}, seurat\_v3 flavor on raw counts, mitochondrial/ribosomal genes excluded), median full-library-size normalized and log1p-transformed, then embedded into 128 principal components (scanpy PCA, zero-centered, fit on all cells) \cite{scanpy}. 
From the cell-level annotations (\texttt{obs}) we derived the three columns. The fine subtype was made location-agnostic by stripping the leading \texttt{Ileal\_}/\texttt{Non\_ileal\_} prefix
from \texttt{minor} (\texttt{minor\_agnostic}), which defines the held-out subtype
\texttt{stem\_LGR5pos}. The tissue axis was collapsed to a binary concept
\texttt{is\_ileum} (\texttt{ileum} if \texttt{Site == Terminal\_Ileum}, else
\texttt{non\_ileum}), since the four colon sub-regions are transcriptomically
hard to distinguish. The final major cell-type concept was obtained by a
deterministic union-merge of the 22-value \texttt{major} column into seven
major lineages (Table~\ref{tab:emily_l1_pub}); the \texttt{Ileal\_epithelium} and
\texttt{Non\_ileal\_epithelium} majors are merged into a single \texttt{epithelial}
lineage so the cell-type concept is not a proxy for \texttt{is\_ileum}.

\begin{table}[t]
  \centering
  \caption{Cell-type concept: mapping of the 22
    \texttt{major} values onto seven lineages.}
  \label{tab:emily_l1_pub}
  \begin{tabularx}{\textwidth}{l >{\raggedright\arraybackslash}X}
    \toprule
    \textbf{Major cell type} & \textbf{Original \texttt{major} values merged} \\
    \midrule
    epithelial   & \texttt{Ileal\_epithelium}, \texttt{Non\_ileal\_epithelium} \\
    T\_ILC       & \texttt{CD4\_T}, \texttt{CD8\_T}, \texttt{Unconventional\_T}, \texttt{Innate\_lymphocytes} \\
    B lineage    & \texttt{B}, \texttt{Plasma} \\
    myeloid cell & \texttt{Mono\_macro}, \texttt{DC}, \texttt{Mast}, \texttt{Cycling\_MNP} \\
    endothelial  & \texttt{Endothelium} \\
    glial cell   & \texttt{Glial} \\
    mesenchymal  & \texttt{Fibroblast}, \texttt{Epi\_fibroblast}, \texttt{LP\_fibroblast}, \texttt{Myofibroblast}, \texttt{C3pos\_fibroblast}, \texttt{THY1pos\_FAPpos\_PDPNpos\_fibroblast}, \texttt{Pericyte}, \texttt{Cycling\_stroma} \\
    \bottomrule
  \end{tabularx}
\end{table}

\subsection{Synthetic data}
\label{sec:sim}

\paragraph{Hierarchical Poisson generative model.}
We adapt the hierarchical Poisson single-cell generator of \cite{Bottleneck}
(\texttt{conceptlab.datagen.OmicsDataGenerator}), reused unchanged as the base model and
extended by the two terms below. Each cell $n$ draws a donor $b_n$, a tissue $t_n$, a cell
type $u_n$ and a binary latent concept vector $v_n\in\{0,1\}^{C}$,
\begin{equation}
  p^{(B)}\!\sim\mathrm{Dir}(10\cdot\mathbf{1}_B),\quad
  p^{(T)}\!\sim\mathrm{Dir}(10\cdot\mathbf{1}_T),\quad
  p^{(U)}_{t}\!\sim\mathrm{Dir}(\mathbf{1}_U),\quad
  p^{(C)}_{u}\!\sim\mathrm{Dir}(\tfrac15\mathbf{1}_C),
\end{equation}
\begin{equation}
  b_n\sim\mathrm{Cat}(p^{(B)}),\quad
  t_n\sim\mathrm{Cat}(p^{(T)}),\quad
  u_n\sim\mathrm{Cat}(p^{(U)}_{t_n}),\quad
  v_{n,c}\sim\mathrm{Bern}\big(\sigma(\zeta_{n,c})\big),\;
  \zeta_n=\mathrm{clip}\big(\mathrm{logit}\,p^{(C)}_{u_n},-8,8\big),
\end{equation}
with gene-wise additive effects
\begin{equation}
  w_g\sim\mathrm{Unif}(\log\ell,\log h),\quad
  \gamma_{b,g}\sim\mathcal{N}(0,\sigma_B^2),\quad
  \omega_{u,g}\sim\mathcal{N}(0,\sigma_U^2),\quad
  \tau_{t,g}\sim\mathcal{N}(0,\sigma_T^2),
\end{equation}
\begin{equation}
  s_{c,g}\sim\mathcal{N}(0,\sigma_C^2)\cdot\mathbf{1}[m_{c,g}=1],\quad
  m_{c,g}\sim\mathrm{Bern}(\pi_c),\quad \pi_c=1-\mathrm{Beta}(a,b),
\end{equation}
a per-cell library offset $\ell_n\sim\mathcal{N}(0,(\sigma^L_{b_n})^2)$,
$\sigma^L_b\sim\mathrm{Unif}(0.01,0.03)$, and gene noise
$\varepsilon_{n,g}\sim\mathcal{N}(0,\sigma_\varepsilon^2)$. The Poisson log-rate is additive,
\begin{equation}
  \log\lambda_{n,g}=w_g+\sum_{c}s_{c,g}v_{n,c}+\omega_{u_n,g}+\gamma_{b_n,g}
  +\tau_{t_n,g}+\ell_n+\varepsilon_{n,g},
  \label{eq:loglambda}
\end{equation}
and counts are drawn by inverse-CDF transform of a stored exogenous uniform
$a_{n,g}\sim\mathrm{Unif}(0,1)$,
\begin{equation}
  x_{n,g}=F^{-1}_{\mathrm{Pois}(\lambda_{n,g})}(a_{n,g}).
  \label{eq:invcdf}
\end{equation}
Keeping $a_{n,g}$ and $\varepsilon_{n,g}$ makes the sampler a deterministic function of the
latents: we call the generator, read the coefficients and exogenous draws back out, and
recompute \eqref{eq:loglambda}--\eqref{eq:invcdf} ourselves, which is what lets us inject the terms below
while holding every other draw fixed. The $T=2$ tissue axis is the editable binary concept
\texttt{is\_ileum}; the model conditions on $\{\texttt{celltype},\texttt{is\_ileum}\}$, while
$v$, the donor fingerprint $\gamma$, the library offset and the noise are unobserved and must
be absorbed by the unknown factors $U$.

\paragraph{Low-rank donor structure.}
In the base model $\gamma_{b,g}$ is i.i.d.\ per gene, so donors are isolated points and a
held-out donor is unpredictable in principle. We replace the donor block by a shared factor
model
\begin{equation}
  \gamma = HW,\qquad
  H\in\mathbb{R}^{B\times R},\ H_{b,r}\sim\mathcal{N}(0,1),\qquad
  W\in\mathbb{R}^{R\times F},\ W_{r,g}\sim\mathcal{N}\!\Big(0,\tfrac{\sigma_B^2}{R}\Big),
  \label{eq:lowrank}
\end{equation}
with the $R$ dense programs $W$ drawn once and shared by all donors. Since
$\mathrm{Var}(\gamma_{b,g})=R\cdot\sigma_B^2/R=\sigma_B^2$, only the \emph{rank} of the donor
signal changes, not its strength; every donor, held-out ones included, lies on the same
$R$-dimensional subspace, so a donor readout learned on training donors transfers.

\paragraph{Nonlinear (heteroskedastic) counterfactual axis and matched linear control.}
The two members are generated from the same seed, hence share all coefficients, per-cell
latents and exogenous draws, and differ only on a block $V$ of the $|V|$ highest-baseline
genes. In the \textbf{linear} member every gene keeps its additive tissue effect, so the
\texttt{is\_ileum} edit is the fixed shift $\tau_{1,\cdot}-\tau_{0,\cdot}$, identical for all
donors. In the \textbf{nonlinear} member the tissue main effect on $V$ is removed
($\tau_{t,g}=0$, $g\in V$) and replaced by a donor--tissue variance interaction
\begin{equation}
  \log\lambda_{n,g}\;\mathrel{+}=\;\rho\,\varphi_{b_n}\,m(t_n),\qquad g\in V,
  \qquad
  \varphi_b\sim\mathcal{N}(0,1)\ \text{with}\ \textstyle\sum_b\varphi_b=0,
  \qquad
  m(0)=1,\ m(1)=M,
  \label{eq:hetero}
\end{equation}
i.e.\ each donor has its own signature $\varphi_b$ on $V$ whose amplitude is multiplied by $M$
in the ileum. By construction
\begin{equation}
  \mathbb{E}[V\mid t]=\rho\,m(t)\,\mathbb{E}[\varphi]=0
  \;\Rightarrow\;
  \mathrm{Cov}(\texttt{is\_ileum},V)=0,
  \qquad
  \frac{\mathrm{Var}[V\mid \text{ileum}]}{\mathrm{Var}[V\mid \text{non\_ileum}]}=M^2 ,
\end{equation}
so the channel is invisible to a covariance penalty but visible to a kernel one: the means
agree in both conditions and the dependence is entirely second-order. It also keeps the
held-out-donor counterfactual well posed, since $\varphi_b$ is present in both conditions
(amplitude $\rho$ at the source, $\rho M$ in the ileum) and is therefore readable from a
donor's source cells; the correct edit is to amplify each donor's own baseline signature by
$M$, whereas storing the raw source amplitude in $U$ under-scales it. A CPU oracle verifies all
three properties before training (measured $\mathrm{Cov}\approx0$, variance ratio $\approx M^2$,
and per-donor Pearson $r\ge0.75$ between the source $V$-mean and the true $\varphi_b$). Both
members are then passed through the identical pipeline used for the real data, median
library-size normalization, $\log(1+x)$, PCA to $128$ components, and evaluated under a
donor holdout: $20$ of the $150$ donors are withheld from training entirely, and the
\texttt{is\_ileum} edit is scored per cell type on those unseen donors.

\paragraph{Final parameters used.}
All parameters finally used for the simulation are listed in Table \ref{tab:sim_hparams}.
\begin{table}[h]
  \centering
  \begin{tabular}{llr}
    \toprule
    \textbf{Symbol} & \textbf{Meaning} & \textbf{Value} \\
    \midrule
    $N$ & cells & $240{,}000$ \\
    $F$ & genes & $3{,}000$ \\
    $B$ & donors (20 held out) & $150$ \\
    $T$ & tissues (\texttt{is\_ileum}) & $2$ \\
    $U$ & cell types & $4$ \\
    $C$ & latent concepts & $3$ \\
    $[\ell,h]$ & baseline-expression range & $[1,\,6]$ \\
    $\sigma_B$ & donor s.d. & $0.30$ \\
    $\sigma_U$ & cell-type s.d. & $0.20$ \\
    $\sigma_T$ & tissue s.d. & $0.35$ \\
    $\sigma_C$ & concept s.d. & $0.05$ \\
    $\sigma_\varepsilon$ & gene-noise s.d. & $0.02$ \\
    $\sigma^{L}$ & library-size s.d. & $\mathrm{Unif}(0.01,0.03)$ \\
    $(a,b)$ & zero-inflation Beta parameters & $(1,\,0.5)$ \\
    $R$ & donor factor rank & $12$ \\
    $|V|$ & variance-block genes & $150$ \\
    $\rho$ & variance-channel base scale & $0.5$ \\
    $M$ & ileum variance multiplier & $3$ \\
    -- & PCA components & $128$ \\
    -- & seed & $0$ \\
    \bottomrule
  \end{tabular}
  \caption{Generator settings for the matched linear / heteroskedastic pair.}
  \label{tab:sim_hparams}
\end{table}

\subsection{Normalized Hilbert Schmidt Independence Criterion}
\label{sec:nhsic}
\paragraph{The nHSIC orthogonality loss.}
HSIC as a differentiable penalty inside a single-cell VAE follows~\cite{lopez2018};
we use the normalized, centered-kernel-alignment form and apply it between the two
bottleneck blocks rather than between the latent and a nuisance variable.
The bottleneck of cell $i$ splits into a concept vector $c_i\in\mathbb{R}^{d_c}$ and an
unknown-factor vector $u_i\in\mathbb{R}^{d_u}$. The two blocks have different dimensions and cannot be compared directly; instead we compare their \emph{pairwise-similarity patterns} over a minibatch of $n$ cells. With an RBF kernel
$k(a,b)=\exp(-\lVert a-b\rVert^2/2\sigma^2)$ this gives two $n\times n$ Gram matrices
\begin{equation}
  K_{ij}=\exp\!\Big(\!-\frac{\lVert c_i-c_j\rVert^2}{2\sigma_c^2}\Big),
  \qquad
  L_{ij}=\exp\!\Big(\!-\frac{\lVert u_i-u_j\rVert^2}{2\sigma_u^2}\Big),
  \qquad i,j=1,\dots,n,
\end{equation}
where entry $ij$ is the similarity of cells $i$ and $j$ in the respective channel ($1$ =
identical, $0$ = far apart). The bandwidths use the median heuristic,
$\sigma^2=\mathrm{median}\{\lVert x_i-x_j\rVert^2\}_{i\neq j}$, evaluated per batch and
\emph{detached} from the computational graph, so the optimizer cannot trivialize the penalty
by inflating $\sigma$ until all similarities collapse to $1$. Both matrices are centered with
\begin{equation}
  H=I_n-\tfrac{1}{n}\mathbf{1}\mathbf{1}^\top,
\end{equation}
which subtracts row, column and grand means in the implicit feature space, so dependence is
judged from deviations rather than from the overall similarity level. Alignment of the two
centered patterns is measured by the (biased) Hilbert-Schmidt Independence Criterion \cite{HSIC}.
\begin{equation}
  \mathrm{HSIC}(c,u)=\frac{1}{(n-1)^2}\operatorname{tr}(KHLH),
  \qquad
  \operatorname{tr}(KHLH)=\sum_{i,j}(HKH)_{ij}(HLH)_{ij},
\end{equation}
i.e.\ an entrywise inner product of the two centered similarity tables: large when cells that
are alike in concept space are also alike in unknown space, zero iff $c$ and $u$ are
independent in the RKHS (nonlinear dependence included, unlike a cross-covariance penalty).
Its magnitude, however, depends on the feature scales and on $n$. Normalizing by the geometric
mean of the two self-alignments gives the centered kernel alignment we use as the orthogonal loss,
\begin{equation}
  \boxed{\;
  \mathcal{L}_\perp( c,u) =
  \mathcal{L}_{\mathrm{nHSIC}}(c,u)=
  \frac{\operatorname{tr}(KHLH)}
       {\sqrt{\operatorname{tr}(KHKH)\cdot\operatorname{tr}(LHLH)}}\;\in[0,1],}
\end{equation}
with $0$ = independent and $1$ = fully dependent. The $(n-1)^{-2}$ factors cancel, so the
value is scale-invariant and comparable across batches, and its weight in the objective is on
an interpretable scale ($\lambda_\perp\approx1$--$10$).

\paragraph{Numerical details.}
Three guards make the criterion differentiable in practice. (i) Squared distances are expanded
as $D_{ij}=\max(\lVert x_i\rVert^2+\lVert x_j\rVert^2-2x_i^\top x_j,\,0)$ rather than obtained
from a Euclidean-distance primitive, whose gradient is undefined at coincident rows (a batch of
near-identical concept predictions would emit NaNs); the $\max(\cdot,0)$ only absorbs
floating-point round-off. (ii) The traces are evaluated as
$\operatorname{tr}(AHBH)=\sum_{ij}(AH)_{ij}(BH)_{ji}$, avoiding the full matrix products.
(iii) The denominator is clamped \emph{inside} the square root,
$\sqrt{\max(\operatorname{tr}(KHKH)\operatorname{tr}(LHLH),\,\epsilon)}$ with
$\epsilon=10^{-12}$: if the concept predictions collapse, $\operatorname{tr}(KHKH)\to0$ and
$\tfrac{d}{ds}\sqrt{s}$ diverges, so clamping the argument bounds the backward pass at
$\tfrac12\epsilon^{-1/2}$ while leaving non-degenerate batches unchanged. The loss returns $0$
for $n<4$, where the estimator is meaningless. Cost is $O(n^2)$ per batch.

\subsection{Model and training}
\label{sec:model}

\paragraph{Architecture.}
We use the concept-bottleneck VAE of \cite{Bottleneck} unchanged, and modify only the
orthogonality penalty (Sec.~\ref{sec:nhsic}). An encoder maps a cell $x\in\mathbb{R}^{d}$ to
$(\mu,\log\sigma^2)$ through a linear embedding into $\mathbb{R}^{1024}$ followed by $4$
residual blocks (\textsc{Linear}$\to$\textsc{LayerNorm}$\to$\textsc{ReLU}$\to$\textsc{Dropout},
$p=0.1$, plus identity skip); $z\in\mathbb{R}^{128}$ is drawn by the reparametrization trick \cite{vae}.
The bottleneck splits $z$ into two heads,
\begin{equation}
  \hat c=\mathrm{sigmoid}(W_c z+b_c)\in[0,1]^{d_c},
  \qquad
  u=\mathrm{ReLU}(W_u z+b_u)\in\mathbb{R}^{d_u}_{\ge0},
  \qquad d_u=128,
\end{equation}
where $\hat c$ are the supervised concepts and $u$ the unknown factors that must absorb
everything else (donor identity, library size, unmodelled state). The skip decoder embeds
$\hat c$ and $u$ separately into $\mathbb{R}^{1024}$ and passes the $u$-embedding through $4$
residual blocks, each of which re-concatenates the concept embedding before its linear layer,
so the concept vector is re-injected at every depth; a final linear layer returns
$\hat x\in\mathbb{R}^{d}$. During training the decoder is conditioned on the \emph{predicted}
$\hat c$ (never on the ground-truth one-hot), so the concept pathway is the only route by
which concept information can reach the reconstruction. A counterfactual replaces only the
edited concept block by its target one-hot, and decodes from the same $u$: abduction of $u$, action on
$c$, prediction.

\paragraph{Objective.}
With a minibatch of $n$ cells,
\begin{equation}
  \mathcal{L}
  =\underbrace{\mathrm{MSE}(\hat x,x)}_{\text{reconstruction}}
  +\beta\,\underbrace{\Big[-\tfrac12\overline{\big(1+\log\sigma^2-\mu^2-\sigma^2\big)}\Big]}_{\mathrm{KL}}
  +\lambda_c\,\underbrace{d_c\cdot\mathrm{BCE}\big(\mathrm{clip}(\hat c,0,1),\,c\big)}_{\text{concept}}
  +\lambda_\perp\,\underbrace{\mathcal{L}_\perp(\hat c,u)}_{\text{orthogonality}} ,
\end{equation}
all terms averaged over cells and coordinates ($\overline{\cdot}$ denotes the mean over the
batch and latent dimensions). The clip to $[0,1]$ guards the BCE against sigmoid round-off.
Concepts are one-hot encoded and scored per column, so a multi-class concept contributes one
BCE term per level. The orthogonality term is either the linear cross-covariance penalty
\begin{equation}
  \mathcal{L}_{\mathrm{cov}}(\hat c,u)=\big\lVert \widehat{\mathrm{Cov}}(u,\hat c)\big\rVert_F^2,
  \qquad
  \widehat{\mathrm{Cov}}(u,\hat c)=\tfrac{1}{n-1}\,\tilde u^\top\tilde c
\end{equation}
($\tilde{\cdot}$ = batch-centered), our normalized-HSIC penalty
$\mathcal{L}_{\mathrm{nHSIC}}$ (Sec.~\ref{sec:nhsic}), or their sum
$\mathcal{L}_{\mathrm{cov}}+\mathcal{L}_{\mathrm{nHSIC}}$ (\texttt{cov\_nhsic}); a single
weight $\lambda_\perp$ multiplies whichever is selected.

\paragraph{Class-balanced concept loss.}
The binary tissue concept is strongly imbalanced ($83\%$ non-ileum) in the Taurus dataset, and an unweighted BCE
lets the head reach a high accuracy while under-fitting the minority level, which is exactly the
concept we edit. We therefore rescale the positive entries of column $j$ by
$w_j=(1-p_j)/p_j$, with $p_j$ the training prevalence of that level, keeping
\texttt{reduction='mean'} so the loss scale of the other concepts is unchanged. For the
\texttt{taurus} runs this gives $w=4.88$ on \texttt{is\_ileum}; the cell-type concept
is left unweighted. The synthetic runs use the unweighted loss (all levels are near-balanced
by construction).

\paragraph{Inputs and concepts.}
Both settings enter the model in the same representation: median library-size normalization,
$\log(1+x)$, the top $3{,}000$ highly variable genes (\texttt{seurat\_v3}), and PCA to $128$
components fitted once on all cells, so $d=128$. This means the whole model training and evaluations run in PCA space. Concepts are supervised one-hot blocks:
\texttt{taurus} uses the $7$-way major lineage label plus the
binary \texttt{is\_ileum} ($d_c=8$); the simulation uses the $4$-way \texttt{celltype} plus the
binary \texttt{is\_ileum} ($d_c=5$). Donor identity and sub cell type are never concepts, but must be recovered
in $u$, which is what the counterfactual evaluation tests.

\paragraph{Splits.}
Two nested splits are used, with different purposes.
\emph{(i) Evaluation holdout}: Removed from training entirely, never seen in any form.
For \texttt{taurus} this is a \emph{global subtype holdout}: all cells of the held-out subtype
in the target condition (\texttt{is\_ileum}~$=$~ileum $\times$ \texttt{stem\_LGR5pos}) are
withheld, $4{,}100$ of $987{,}743$ cells, leaving $983{,}643$ for training. At evaluation the
retained Rectum \texttt{stem\_LGR5pos} cells are edited non\_ileum~$\to$~ileum and compared to
the real held-out ileal cells, per donor for the $14$ donors that carry the subtype on both
sides, and pooled. For the simulation this is a \emph{donor holdout}: $20$ of the $150$ donors
are withheld in full ($207{,}231$ of $240{,}000$ cells remain), and the edit is scored per cell
type on donors the model has never seen (Sec.~\ref{sec:sim}).
\emph{(ii) Train/validation}: A uniformly random $10\%$ of the remaining cells is held out
as a validation set for monitoring, early stopping and checkpoint selection only:
$885{,}279$~/~$98{,}364$ for \texttt{taurus} and $186{,}508$~/~$20{,}723$ for the simulation.
This split is disjoint from (i) and is drawn from a fixed seed, so training and inference
reconstruct identical splits from the observation table without caching them.

\paragraph{Optimization.}
Adam \cite{adam}, learning rate $3\times10^{-4}$ with an exponential per-epoch decay
($\gamma=0.997$), batch size $512$ with reshuffling each epoch, at most $256$ epochs. Training
stops early when the total validation loss has not improved for \texttt{patience} consecutive
epochs (\texttt{min\_delta}~$=0$), and the checkpoint with the lowest total validation loss, 
not the last one, is the model used for every reported evaluation. Gradients are clipped to
a global norm of $1.0$ in the synthetic runs and left unclipped in the \texttt{taurus} runs.
All runs use seed $0$ for the data split, weight initialization, dropout and batch order.

\paragraph{Training diagnostics.}
Two quantities are logged every $10$ epochs, since the scalar concept loss hides the two
failure modes we care about. \emph{Per-concept accuracy} (argmax within a one-hot block,
threshold $0.5$ for binary concepts) is compared against the majority-class rate of that
concept, detecting a collapsed concept head. The \emph{edit ratio} measures whether edits reach
the output: validation cells are encoded once and decoded twice from the same $u$, with the
edited concept block set to source and to target,
\begin{equation}
\mathrm{edit\ ratio}
=\frac{\big\lVert \mathrm{dec}(c^{\text{tgt}},u)-\mathrm{dec}(c^{\text{src}},u)\big\rVert_2}
        {\big\lVert \mathrm{dec}(c^{\text{src}},u)\big\rVert_2}.
\end{equation}
A ratio near zero means the decoder reconstructs from $u$ alone and ignores the concept, so no
counterfactual can propagate regardless of concept accuracy.

\paragraph{Configurations.}
Table~\ref{tab:train_hparams} lists the settings that differ between runs; everything else is
shared. On the real data we report the selected model (\texttt{cov\,+\,nHSIC},
$\lambda_\perp=10$) together with the two covariance-only baselines at $\lambda_\perp=0.5$ and
$\lambda_\perp=10$, which isolate the penalty type from its strength. The baselines were run
with the more generous stopping patience ($50$ vs.\ $25$) and still converged and stopped much
earlier ($87$ and $71$ epochs, best epochs $37$ and $21$) than the selected model ($173$
epochs, best epoch $148$), so their weaker edits are not a consequence of a shorter budget. On
the simulation we train the full grid $\{\text{cov},\ \text{nHSIC},\ \text{cov\,+\,nHSIC}\}
\times\{0.5,1,5\}$ on both members of the matched pair ($18$ runs) with a single shared
configuration on NVIDIA A100 GPUs.

\begin{table}[h]
  \centering
  \small
  \begin{tabular}{lccc}
    \toprule
    & \multicolumn{2}{c}{\texttt{taurus} (Rectum $\to$ ileum, stem)} & Synthetic pair \\
    \cmidrule(lr){2-3}\cmidrule(lr){4-4}
    & cov baselines & selected & grid \\
    \midrule
    Orthogonality penalty      & cov            & cov\,+\,nHSIC & cov / nHSIC / cov\,+\,nHSIC \\
    $\lambda_\perp$            & $0.5$, $10$    & $10$          & $0.5$, $1$, $5$ \\
    Class-balanced concepts    & \texttt{is\_ileum} & \texttt{is\_ileum} & none \\
    Patience                   & $50$           & $25$          & $50$ \\
    Gradient clipping          & --             & --            & $1.0$ \\
    Epochs run (best epoch)    & $71\,(21)$, $87\,(37)$ & $173\,(148)$ & up to $256$ \\
    \midrule
    \multicolumn{4}{l}{\emph{Shared:}\; $d=128$ (PCA), latent $128$, $d_u=128$, hidden $1024$,
      $4$ layers, dropout $0.1$, skip decoder,} \\
    \multicolumn{4}{l}{\phantom{\emph{Shared:}}\; $\beta=10^{-5}$, $\lambda_c=0.1$, Adam
      $3\times10^{-4}$, $\gamma_{\mathrm{lr}}=0.997$, batch $512$, max $256$ epochs, seed $0$,
      val.\ fraction $0.1$.} \\
    \bottomrule
  \end{tabular}
  \caption{Training configurations. All models share one architecture and optimizer; the runs
  differ only in the orthogonality penalty and its weight (plus the stopping patience and
  gradient clipping noted above).}
  \label{tab:train_hparams}
\end{table}
\subsection{Evaluation}
\label{sec:eval}
\paragraph{Distances.}
Two distances between two sets of cells are used, both computed in the
$128$-dimensional PCA space the model is trained in. The first is the unbiased
squared maximum mean discrepancy, written
$\mathrm{MMD}$ throughout, with a Gaussian kernel,
\begin{equation}
  \mathrm{MMD}(X,Y)=
  \frac{1}{n(n-1)}\sum_{i\neq j}k(x_i,x_j)
  +\frac{1}{m(m-1)}\sum_{i\neq j}k(y_i,y_j)
  -\frac{2}{nm}\sum_{i,j}k(x_i,y_j),
\end{equation}
with $k(a,b)=\exp(-\varsigma\lVert a-b\rVert^2)$ and the bandwidth set by the
median heuristic on the pooled sample, $\varsigma=1/(2\,\mathrm{median}\{\lVert
z_i-z_j\rVert\}^2)$, $z=X\cup Y$. Self-similarity terms are excluded and the
estimate is clipped at $0$. The second is the debiased Sinkhorn divergence
$S(X,Y)=\mathrm{OT}_\epsilon(X,Y)-\tfrac12\mathrm{OT}_\epsilon(X,X)
-\tfrac12\mathrm{OT}_\epsilon(Y,Y)$ with entropic regularization
$\epsilon=0.1$ on squared-Euclidean cost matrices, each normalized by its own
maximum before solving and the divergence rescaled back. Both groups are
subsampled to at most $2{,}000$ cells with a fixed seed before either distance is
computed.

\paragraph{Ratio metrics.}
Absolute distances are not comparable across subtypes or datasets, so, following
\cite{Bottleneck}, we normalize by the distance from the target to the closest
population that already exists in the training data. Writing $\hat{X}$ for the
edited cells and $X^{\mathrm{tgt}}$ for the real withheld target cells,
\begin{equation}
  \mathrm{rMMD}=\frac{\mathrm{MMD}(\hat{X},X^{\mathrm{tgt}})}
  {\min_{p\in\mathcal{P}}\mathrm{MMD}(X_p,X^{\mathrm{tgt}})},
  \qquad
  \mathrm{rSinkhorn}=\frac{S(\hat{X},X^{\mathrm{tgt}})}
  {\min_{p\in\mathcal{P}}S(X_p,X^{\mathrm{tgt}})},
  \label{eq:rmmd}
\end{equation}
where $\mathcal{P}$ is a set of candidate populations drawn from the
\emph{training} cells only. A value below $1$ means the edit is a better
description of the target than the best available "map the cells onto the
nearest existing population" answer; the numerator, the identity of the winning
population $p^\star$ and the full candidate list are recorded for every run. The
minimum is taken separately for each metric, so the winner may differ between
rMMD and rSinkhorn.

\paragraph{Candidate populations, pooled subtype holdout (Taurus).}
Here $\mathcal{P}$ is the set of joint combinations of the model's own concept
columns: every observed pair (major lineage, \texttt{is\_ileum}) among the
training cells, with populations of fewer than $10$ cells dropped. For the
Taurus runs this yields $14$ candidates from the $7$ lineages crossed with the
binary tissue concept. Fine subtypes are deliberately \emph{not} used to define
candidates, so the baseline is "the closest population the model is allowed to
name", not "the same subtype in the target condition", which would not exist for
a held-out subtype.
For all three Taurus runs the winning population was
(epithelial, non\_ileum) at MMD $0.244$, i.e.\ the un-edited source lineage, and
the same population also won under the Sinkhorn divergence.

\paragraph{Candidate populations, donor holdout (simulation).}
The donor holdout scores each (held-out donor $b$, cell type $k$) pair
separately. The numerator is the distance between the edit of that donor's
source cells of type $k$ and the same donor's real target cells of type $k$. The
denominator is the distance between the \emph{generic} training population of the
target condition and cell type $k$, pooled over all training donors, and that
same donor's real target cells. The ratio therefore isolates donor specificity:
a model that predicts the condition correctly but ignores who the donor is scores
$\approx 1$ by construction, and only a model that recovers the donor's own
position scores below it. A (donor, cell type) row is scored only if both sides
have at least $20$ cells and the denominator population has at least $10$; we
report the median over the surviving rows and, where stated, the fraction of rows
below $1$. For the final simulation pair this gives $65$ rows across the $20$
held-out donors. %

\paragraph{Subtype screen.}
The held-out subtype for the pooled evaluation was chosen before any training by
a model-free screen. For each of the $36$ location-agnostic fine subtypes we
computed the real source-versus-target distance (MMD and Sinkhorn, per donor and
pooled) together with the number of donors carrying the subtype on both sides and
the cell counts per side. Subtypes whose two conditions barely differ are
unsuitable, because the denominator of Eq.~\ref{eq:rmmd} then collapses onto the
source population and the ratio becomes unstable. \texttt{stem\_LGR5pos} ranked
third by per-donor MMD ($0.201$; pooled $0.146$) with $14$ paired donors,
$12{,}614$ source and $4{,}100$ target cells, and was preferred over the two
larger-gap subtypes for its higher paired-donor count and its coherence in the
target condition. Because the holdout is baked into training, each candidate
subtype would require its own model, so the screen is run first and only the
selected subtype's hold-out model is trained.

\subsection{Use of LLMs.}
During the preparation of this work, the authors used Claude Opus 4.8, Opus 5 and Opus 5.5 to improve the code, language, readability, and flow of the manuscript. After using this tool, the authors reviewed and edited the content as needed and take full responsibility for the scientific integrity and final content of the publication.

\printbibliography %

\end{document}